\documentclass[a4paper,fleqn]{cas-dc}

\usepackage{lineno,hyperref}
\usepackage{times}
\usepackage{epsfig}
\usepackage{graphicx}
\usepackage{amsmath}
\usepackage{amssymb}

\usepackage{caption}
\usepackage{xcolor}

\usepackage{bm}
\usepackage{microtype}

\usepackage{amsthm}

\usepackage{subfigure} 
\usepackage{comment}
\usepackage{algorithm}
\usepackage{algpseudocode} 
\usepackage{multirow}

\usepackage{color}
\usepackage{hyperref}

\usepackage[numbers]{natbib}

\def\tsc#1{\csdef{#1}{\textsc{\lowercase{#1}}\xspace}}
\tsc{WGM}
\tsc{QE}

\begin{document}
\let\WriteBookmarks\relax
\def\floatpagepagefraction{1}
\def\textpagefraction{.001}

\shorttitle{Source-free Domain Adaptation by casting a BAIT}

\shortauthors{Shiqi Yang, Yaxing Wang, Luis Herranz, Shangling Jui, Joost van de Weijer}  

\title [mode = title]{Casting a BAIT for Offline and Online Source-free Domain Adaptation}  



%

\author[1,2]{Shiqi Yang}[]
\author[3]{Yaxing Wang}[]
\cormark[1]
\cortext[1]{Corresponding author: yaxing@nankai.edu}

\author[1,2]{Luis Herranz}[]
\author[4]{Shangling Jui}[]
\author[1,2]{Joost van de Weijer}[]


\affiliation[1]{organization={Computer Vision Center},
            city={Bellaterra},
            country={Spain}}

\affiliation[2]{organization={Department of Computer Science},
            addressline={Universitat Autònoma de Barcelona}, 
            city={Bellaterra},
            country={Spain}}
            
\affiliation[3]{organization={Nankai University},
            city={Tianjin},
            country={China}}
            
\affiliation[4]{organization={Huawei Kirin Solution},
            city={Shanghai},
            country={China}}









\begin{abstract}
We address the source-free domain adaptation (SFDA) problem, where only the source model is available during adaptation to the target domain. We consider two settings: the offline setting where all target data can be visited multiple times (epochs) to arrive at a prediction for each target sample, and the online setting where the target data needs to be directly classified upon arrival. Inspired by diverse classifier based domain adaptation methods, in this paper we introduce a second classifier, but with another classifier head fixed. When adapting to the target domain, the additional classifier initialized from source classifier is expected to find misclassified features. Next, when updating the feature extractor,  those features will be pushed towards the right side of the source decision boundary, thus achieving source-free domain adaptation. Experimental results show that the proposed method achieves competitive results for offline SFDA on several benchmark datasets compared with existing DA and SFDA methods, and our method surpasses by a large margin other SFDA methods under online source-free domain adaptation setting.
\end{abstract}




\maketitle


\section{Introduction}

Though achieving great success, typically deep neural networks demand a huge amount of labeled data for training. However, collecting labeled data is often laborious and expensive. It would, therefore, be ideal if the knowledge obtained on label-rich datasets can be transferred to unlabeled data. For example, after training on synthetic images, it would be beneficial to transfer the obtained knowledge to the domain of real-world images. However, deep networks are weak at generalizing to unseen domains, even when the differences are only subtle between the datasets~\cite{oquab2014learning}. 
In real-world situations, a typical factor impairing the model generalization ability is the distribution shift between data from different domains.

\textit{Domain Adaptation} (DA)~\cite{long2015learning,tzeng2017adversarial,cicek2019unsupervised,mozafari2017cluster,wang2022informative,wang2021interbn,wang2023boosting,wang2023reducing} aims to reduce the domain shift between labeled and unlabeled target domains. Early works~\cite{gong2012geodesic} learn domain-invariant features to link the target domain to the source domain. Along with the growing popularity of deep learning, many works benefit from its powerful representation learning ability for domain adaptation. Those methods typically minimize the distribution discrepancy between two domains~\cite{long2018transferable}, or deploy adversarial training~\cite{zhang2019domain,lu2020stochastic}.

However, a crucial requirement in the methodology of these methods is that they require access to the source domain data during the adaptation process to the target domain. Accessibility to the source data of a trained source model is often impossible in many real-world applications, for example deploying domain adaptation algorithms on mobile devices where the computation capacity is limited, or in situations where data-privacy rules limit access to the source domain. Without access to the source domain data, the above methods suffer from inferior performance.

Because of its relevance and practical interest, the \textit{source-free adaptation} (SFDA) setting where the model is first trained on the source domain and has no longer access to the source data afterward, has started to get traction recently~\cite{kundu2020universal, liang2020we, li2020model}. In this paper, we further distinguish between offline and online SFDA. In the offline case, the algorithm can access the target data several times (or epochs) before arriving at a class prediction for each of the samples in the target data. In the online (or streaming) case, the algorithm has to directly predict the label of the incoming target data, meaning that there is only a single pass (or epoch) over the target data. The online scenario is often more realistic, since often an algorithm is expected to directly perform when being exposed to a new domain (as is common in for example robotics applications) and cannot wait with its prediction until it has seen all target data.

Existing method in SFDA have focused on offline SFDA. Among these methods, USFDA~\cite{kundu2020universal} addresses universal DA~\cite{you2019universal} and SF~\cite{kundu2020towards} addresses for open-set DA~\cite{saito2018open}. Both have the drawback of requiring to generate images or features of non-existing categories.  SHOT~\cite{liang2020we} and 3C-GAN~\cite{li2020model} address close-set SFDA. 3C-GAN~\cite{li2020model} is based on target-style image generation by a conditional GAN, which demands a large computation capacity and is time-consuming. Meanwhile, SHOT proposes to transfer the source hypothesis, i.e. the fixed source classifier, to the target data.  {Also, the pseudo-labeling strategy is an important step of the SHOT method.}  
However, SHOT has two limitations. First, it needs to access all target data to compute the pseudo labels, only after this phase it can start adaptation to the target domain. This is infeasible for online streaming applications 
 where the system is expected to directly process the target data and data cannot be revisited. Secondly, it heavily depends on pseudo-labels being correct. Therefore, some wrong pseudo-labels may compromise the adaptation process.

Our method is inspired by the diverse classifiers based DA method MCD~\cite{saito2018maximum}. However, that work fails for SFDA. Like that work we also deploy two classifiers to align target with source classifier. 
In our method, after getting the source model, we propose to freeze the classifier head of the source model during the whole adaptation process. The decision boundary of this source classifier serves as an anchor for SFDA. Next, we add an extra classifier (called \textit{bait} classifier) initialized from the source classifier (referred to as \textit{anchor} classifier). The bait classifier is expected to find those target features that are misclassified by the source classifier.
By encouraging the two classifiers to have similar predictions, the feature extractor will push target features to the correct side of the source decision boundary, thus achieving adaptation. In the experiments, we show that our method, dubbed BAIT, achieves competitive results compared with methods using source data and also other SFDA methods.  {Moreover, other than SHOT, our method can directly start adaptation to the target domain when target data arrives, and does not require a full pass through the target data before starting adaptation. As a consequence, our method obtains superior results in the more realistic setting of online source-free domain adaptation.} 


We summarize our contributions as follows:
\begin{itemize}
	\item We propose a new method for the challenging source-free domain adaptation scenario. under either online or offline setting. Our method does neither require image generation as in \cite{li2020model,kundu2020universal,kundu2020towards} and does not require the usage of pseudo-labeling~\cite{liang2020we}. 
	\item Our method prevents the need for source data by deploying an additional classifier to align target features with the source classifier. We thus show that the previously popular diverse classifiers methods designed for DA (\cite{saito2018maximum}) can be extended to SFDA by introducing a fixed classifier, entropy based splitting and a class-balance loss.
	\item We demonstrate that the proposed BAIT approach obtains similar results or outperforms existing DA and SFDA methods on several datasets.  
\end{itemize}

\section{Related Works}
\noindent \textbf{Domain adaptation with source data}. Early moment matching methods align feature distributions by minimizing the feature distribution discrepancy, including methods such as DAN~\cite{long2015learning} which deploys Maximum Mean Discrepancy. CORAL~\cite{sun2016return} matches the second-order statistics of the source to target domain. Inspired by adversarial learning, 
CDAN~\cite{long2018conditional} trains a deep networks conditioned on several sources of information. DIRT~\cite{shu2018dirt} performs domain adversarial training with an added term that penalizes violations of the cluster assumption. Domain adaptation has also been tackled from other perspectives. 
DAMN~\cite{bermudez2020domain} introduces a framework where each domain undergoes a different sequence of operations. 

\textbf{Domain adaptation without source data}. All these methods, however, require access to source data during adaptation. Recently, USFDA~\cite{kundu2020universal} and FS~\cite{kundu2020towards} explore the source-free setting, but they focus on the universal DA task~\cite{you2019universal} and open-set DA~\cite{saito2018open}, where the label spaces of source and target domain are not identical. And their proposed methods are based on generation of simulated negative labeled samples during source straining period, in order to increase the generalization ability for unknown class. Most relevant works are SHOT~\cite{liang2020we} and 3C-GAN~\cite{li2020model}, both are for close-set DA. SHOT needs to compute and update pseudo labels before updating model, which has to access all target data and may also have negative impact on training from the noisy pseudo labels, and 3C-GAN needs to generate target-style training images based on conditional GAN, which demands large computation capacity.
Instead of synthesizing target images or using pseudo labels, our method introduces an additional classifier to achieve feature alignment with the fixed source classifier. {And recently there is also research line addressing SFDA by clustering~\cite{yang2021generalized,yang2021exploiting,yang2022attracting}, however, those methods need to build memory bank for neighbors retrieving, thus they are not feasible for online domain adaptation.} Our work is inspired by MCD~\cite{saito2018maximum}, however, it is more efficient and performs well under the source-free setting. It is important to note that for MCD source supervision is crucial during adaptation on target.

\section{BAIT for Source-Free Domain Adaptation}
We start by introducing our normal offline source-free domain adaptation method. Finally, we will extend this method to the online case where target data are only seen once. 

We denote the labeled source domain data with $n_s$ samples as $\mathcal{D}_s = \{(x_i^s,y^s_i)\}_{i=1}^{n_s}$, where $y^s_i$ is the corresponding label of $x_i^s$, and the unlabeled target domain data with $n_t$ samples as $\mathcal{D}_t=\{x_j^t\}_{j=1}^{n_t}$, and the number of classes is $K$. Usually, DA methods eliminate the domain shift by aligning the feature distribution between the source and target domains. Unlike the normal setting, we consider the more challenging SFDA setting which during adaptation to the target data has no longer access to the source data, and has only access to the model trained on the source data. 

\subsection{Source classifier as anchor}\label{sec:c1}

\begin{figure*}[t]
	\begin{center}
		\includegraphics[width=0.95\textwidth]{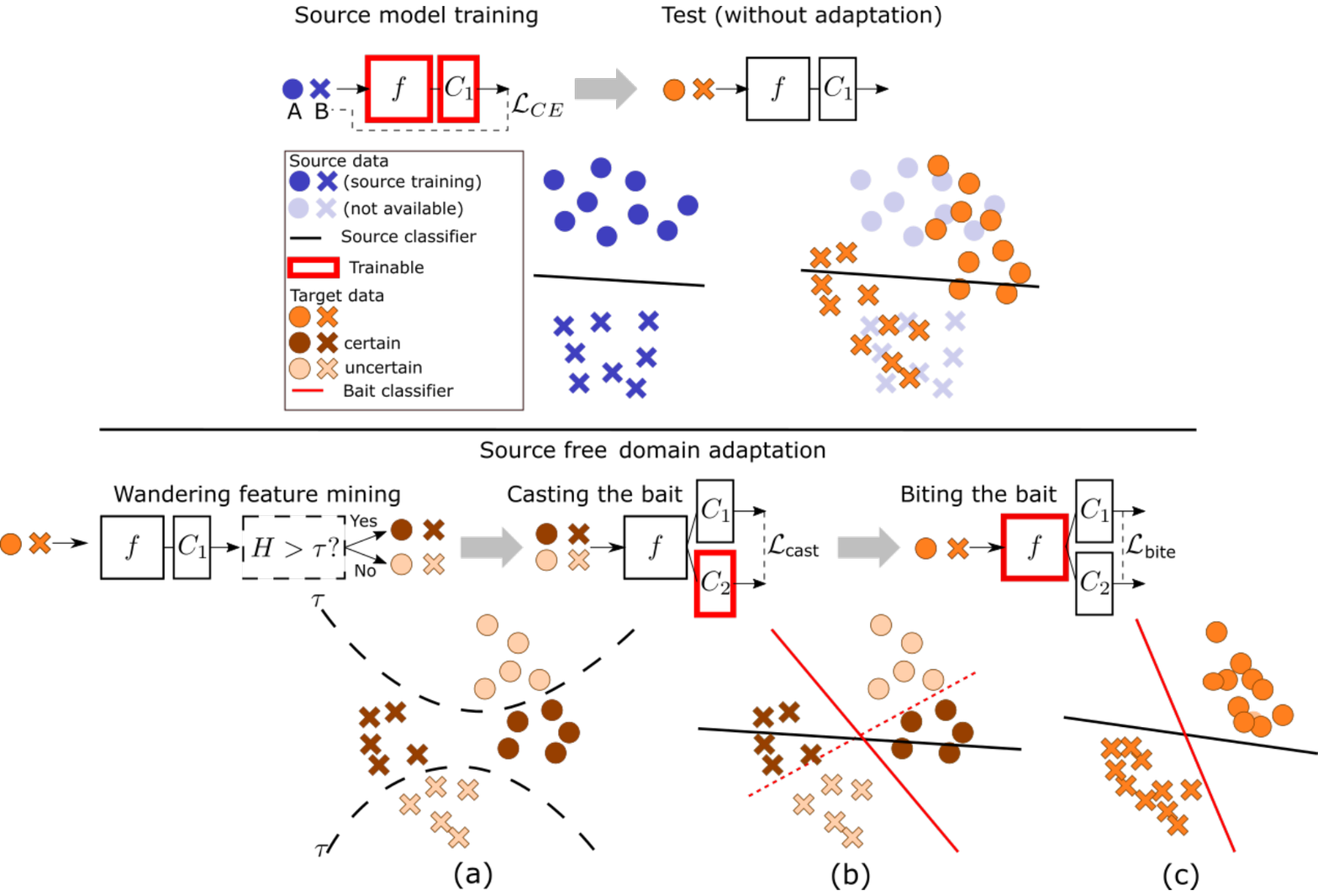}
	\end{center}\vspace{-4mm} 
	\caption{{Illustration of training process. The top shows that the source-training model fails on target domain due to domain shift. The bottom illustrates our adaptation process. (a): splitting feature in \textbf{current batch} into 2 groups by the prediction entropy $H$ and the threshold $\tau$, (b) then increasing prediction divergence between two classifiers for uncertain features but keep the prediction unchanged for uncertain features, meanwhile maximizing KL divergence can also prevent bait classifier from moving to the undesirable position (dashed red line). (c): training the feature extractor pushes all features to the same side of $C_1$ and $C_2$.\vspace{-2mm}}}
	\label{fig:training_policy}
	\vspace{-2mm}
\end{figure*}

We decompose the neural network into two parts: a feature extractor $f$, and a classifier head $C_1$ which only contains one fully connected layer (with weight normalization). We first train the baseline model on the labeled source data $\mathcal{D}_s$ with standard cross-entropy loss:
\begin{equation}\label{eq:ce}
	\small
	\mathcal{L}_{\mathrm{CE}}= -\frac{1}{n_s}\sum_{i=1}^{n_s}\sum_{k=1}^K {I_{[k=y_{i}^s]}} \log p_k(x_i^s)
\end{equation}
where the $p_k$ is the $k$-th element of the softmax output, and $I_{\left [ z \right ]}$ is the indicator function which is 1 if $z$ is true, and 0 otherwise.

A closer look at the training process of DA methods unveils that the feature extractor aims to learn a discriminative representation, and the classifier strives to distinguish the representations of the various classes. DA methods tackle domain shift by aligning the feature distribution (from the feature extractor) of the source and target domains. A successful alignment of the features means that the features produced by the feature extractor $f$ from both domains will be classified correctly by the classifier head.

As shown in Fig~\ref{fig:training_policy}(left), due to the domain shift, the cluster of target features generated by the source-training feature extractor will deviate from the source class prototype, meaning some target features will locate at the wrong side of the source decision boundary. 
Similar to \cite{kundu2020universal,liang2020we}, we freeze the source-trained classifier $C_1$. This implicitly allows us to store the relevant information from the source domain, \textit{i.e.}, the position of the source decision boundary. With the source classifier as an anchor in the feature space, we hope to push the target features towards the right side of the decision boundary. Hereafter, we refer to classifier $C_1$ as the \emph{anchor classifier}.

\subsection{Second classifier as bait}
For the fixed anchor classifier to be successful for source-free domain adaptation, we require to address two problems. First, part of the target data will not be well classified (have uncertain predictions) due to the domain shift, and this data needs to be identified. Secondly, we have to adapt the feature extractor in such a way that this data can subsequently be classified correctly by the anchor classifier. Therefore, we propose the BAIT method that is a two-step algorithm which exactly addresses these two problems. Our method is shown in Fig.~\ref{fig:training_policy}(right). After training the model on the source data, we get a feature extractor $f$, and an anchor classifier $C_1$. We fix $C_1$ in the subsequent training periods, and use it to initialize an extra classifier $C_2$. The extra classifier $C_2$ is optimised to find target features which are not clearly classified by $C_1$. Next, those target features are to be pulled towards the right side of source classifier. Hereafter, we refer to the classifier $C_2$ as the \emph{bait classifier}.

{In order to train the desired $C_2$, we propose a 2-step training policy which alternates between training the bait classifier $C_2$ and feature extractor $f$. It is inspired by diverse classifiers based DA method, such as MCD~\cite{saito2018maximum}. Unlike those methods which train all classifiers along with source data, our method addresses source-free domain adaptation with the fixed anchor classifier and the learnable bait classifier. We experimentally show that the original MCD cannot handle SFDA while our proposed method performs well under this setting.}

\noindent \textbf{Step 1: casting the bait.} In step 1, we only train bait classifier $C_2$, and freeze feature extractor $f$. 
As shown in Fig.~\ref{fig:training_policy}, due to the domain shift, some target features will not locate on the right side of the source decision boundary, which is also referred to as misalignment~\cite{cicek2019unsupervised,jiang2020implicit}. In order to align target features with the source classifier, we use the bait classifier to find the those features at the wrong side of the anchor classifier/decision boundary (uncertain features). 

Therefore, before adaptation, we split the features of the current \textbf{mini-batch} of data into two sets: the uncertain $\mathcal{U}$ and certain set $\mathcal{C}$, as shown in Fig.~\ref{fig:training_policy} (a), according to their prediction entropy:
\begin{equation}\label{eq:wandering_set}
	\begin{aligned}
		\mathcal{U}=\left\{x\vert x\in \mathcal{D}_t,H\left(p^{(1)}\left(x\right)\right)>\tau \right\} ,\\
		\mathcal{C}=\left\{x\vert x\in \mathcal{D}_t,H\left(p^{(1)}\left(x\right)\right)\leq\tau \right\}
	\end{aligned}
\end{equation}
where $p^{(1)}\left(x\right)=\sigma(C_1(f(x))$ is the prediction of the anchor classifier ($\sigma$ represents the softmax operation) and $H(p(x))=-\sum_{i=1}^K p_i \log p_i$.
The threshold $\tau$ is estimated as a percentile of the entropy of $p_1\left(x\right)$ in the mini-batch. We empirically found that choosing $\tau$ such that the data is equally split between the certain and uncertain set provided good results (also see ablation).

Having identified the certain and uncertain features, we now optimize the bait classifier to reduce the symmetric KL divergence for the certain features, while increasing it for the uncertain features. As a consequence, the two classifiers will agree for the certain features but disagree for the uncertain features. This is achieved by following objective:
\begin{equation}\label{eq:casting_bait}
\begin{aligned}
	\mathcal{L}_{\mathrm{cast}}(C_2)=\sum_{x\in{\mathcal{C}}}\mathop{D_{SKL}}(p^{(1)}(x),p^{(2)}(x)) \\- 
	\sum_{x\in\mathcal{U}}\mathop{D_{SKL}}(p^{(1)}(x),p^{(2)}(x))
\end{aligned}
\end{equation}
where $\mathop{D_{SKL}}$ is the symmetric KL divergence: $\mathop{D_{SKL}}(a,b)=\frac{1}{2}\left(\mathop{D_{KL}}(a|b)+\mathop{D_{KL}}(b|a)\right)$. {Note that $KL(p^{(2)}||p^{(1)})=-H(p^{(2)})-\sum p^{(2)}log p^{(1)}$. Instead of using L1 distance like MCD~\cite{saito2018maximum}, the advantage of maximizing KL divergence is that it can prevent the bait classifier from moving to the undesirable place, as the dashed red line shown in the Fig.~\ref{fig:training_policy}(b), since minimizing entropy will encourage the decision boundary not to go across the dense feature region according to cluster assumption~\cite{chapelle2005semi,grandvalet2005semi,shu2018dirt}.}

As shown in Fig.~\ref{fig:training_policy} (a-b), given that $C_2$ is initialized from $C_1$, increasing the KL-divergence on the uncertain set between two classifiers will drive the boundary of $C_2$ to those features with higher entropy. Decreasing it on the certain set encourages the two classifiers to have similar predictions for those features. This will ensure that the features with lower entropy (of high possibility with correct prediction) will stay on the same side of the classifier.

\noindent \textbf{Step 2: biting the bait.} In this stage, we only train the feature extractor $f$, aiming to pull the target features towards the same side of two classifiers.
Specifically, we update the feature extractor by minimizing the proposed bite loss:
\begin{equation}\label{eq:bite}
	\mathcal{L}_{\mathrm{bite}}\left(f\right)=\sum_{i=1}^{n_t}\sum_{k=1}^{K}[-p^{(2)}_{i,k}\log p^{(1)}_{i,k}
	-p^{(1)}_{i,k}\log p^{(2)}_{i,k}]
\end{equation}
By minimizing this loss, the prediction distribution of the bait classifier should be similar to that of the anchor classifier and vice verse, which means target features are excepted to locate on the same sides of the two classifiers.

Intuitively, as shown in Fig.~\ref{fig:training_policy} (c), minimizing the bite loss $\mathcal{L}_{\mathrm{bite}}$ will push target features towards the right direction of the decision boundary. Metaphorically, in this stage the anchor classifier bites the bait (those features with different predictions from anchor and bait classifier) and pushes it towards the anchor classifier. 

Additionally, in order to avoid the degenerate solutions, which allocate all uncertain features to some specific class, we adopt the class-balance loss $\mathcal{L}_b$ to regularize the feature extractor~\cite{ghasedi2017deep,shi2012information}:
\begin{equation}\label{eq:balance}
	\mathcal{L}_{\mathrm{b}}\left(f\right)=\sum_{k=1}^K[KL(\bar{p}_k^{(1)}(x))||{q}_k)+KL(\bar{p}_k^{(2)}(x))||{q}_k)]
\end{equation}
where $\bar{p}_k=\frac{1}{n_t}\sum_{x\in\mathcal{D}_t} p_k(x)$ is the empirical label distribution, and ${q}$ is uniform distribution ${q}_k=\frac{1}{K}, \sum_{k=1}^K {q}_k=1$. With the class-balance loss $\mathcal{L}_b$, the model is expected to have a more balanced prediction.

\noindent \textbf{Online source-free domain adaptation}
\label{sec:online}
As discussed in the introduction, for many applications the current paradigm of offline SFDA is not realistic. This paradigm requires the algorithm to first collect all target data (and be able to process it multiple times) before arriving at a class prediction for each target dataset sample. In the online case, the algorithm has to directly provide class predictions as the target data starts arriving. This scenario is for example typical in robotics applications, where the robot has to directly function when arriving in a new environment.

Our proposed method can be straightforwardly extended to online SFDA. Since in this case the predictions of the fixed classifier have only a low reliability, we found that it was beneficial in the online setting to remove the entropy based splitting. During the adaptation, the target data are only accessible once, \textit{i.e.}, we only train one epoch.


\section{Experiments}
In the following, we first test our method on a toy dataset. Then we provide detailed experiments under offline setting. Finally, we evaluate our method under online setting.

\subsection{Experiment on Twinning moon dataset}\label{sec:toy}
\begin{figure}[tbp]
	\begin{center}
		\includegraphics[width=0.44\textwidth]{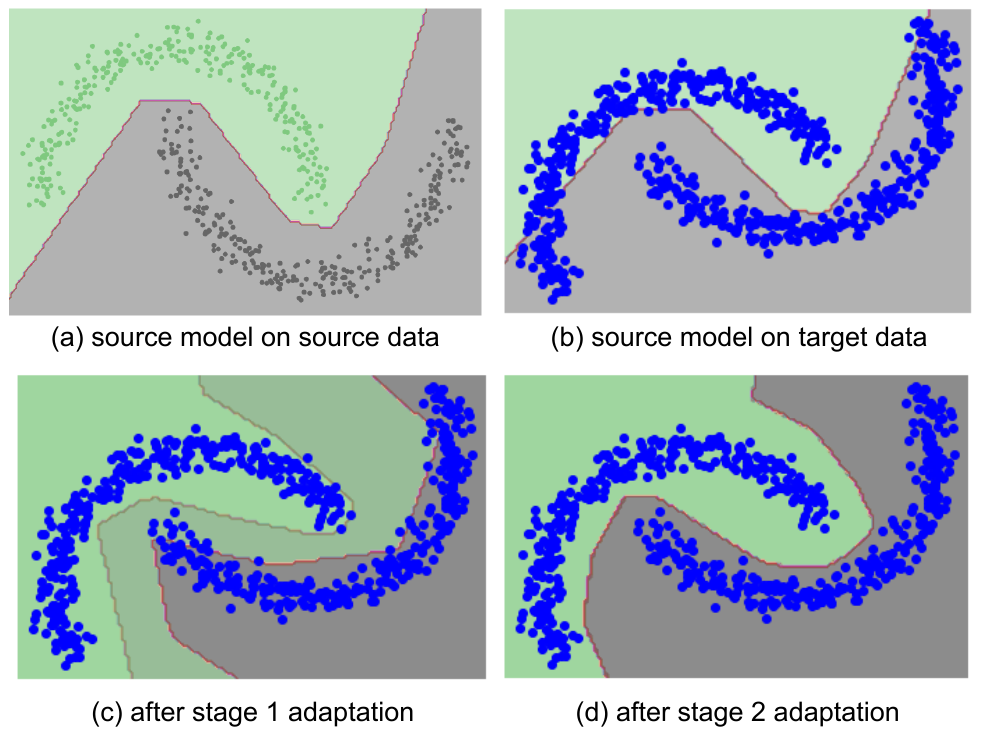}
	\end{center}
 \vspace{-4mm}
	\caption{{Toy experiment on the twinning moon 2D dataset. The blue points refer to target data. The green and gray refer to source data. Decision boundaries after training model only on the source data and testing on source (a) and target (b) data. (c) After stage 1 training in the middle of adaptation with only target data. The two borderlines denote two decision boundaries (with $C_1$ in red). (d) After stage 2 training, the two decision boundaries almost coincide.\vspace{-2mm}}}
	\label{fig:moon}
	\vspace{-2mm}
\end{figure}

We carry out our experiment on the twinning moon dataset. For this data set, the data from the source domain are represented by two inter-twinning moons, which contain 300 samples each. We generate the data in the target domain by rotating the source data by $30^{\circ}$. Here the rotation degree can be regarded as the domain shift. First we train the model only on the source domain, and test the model on all domains. As shown in Fig.~\ref{fig:moon}(a) and (b), due to the domain shift the model performs worse on the target data. Then we adapt the model to the target domain with the anchor and bait classifiers, without access to any source data. As shown in Fig~\ref{fig:moon}(c) during adaptation the bait loss moves the decision boundary\footnote{Note here the decision boundary is from the whole model, since the input are data instead of features.} of the bait classifier to different regions than the anchor classifier. After adaptation the two decision boundaries almost coincide and both classifiers give the correct prediction, as shown in Fig.~\ref{fig:moon}(d).

\begin{table}[t]
\caption{Results for various domain adaptations on Office-31 and VisDA.  {A2D refers to an adaptation from domain A(Amazon) to D(DSLR), etc}. The three methods at the bottom are \textbf{source-free} methods. {'Avg' means average precision over all tasks, while 'Per-class' means average per-class accuracy over 12 classes.}\vspace{-2mm}}\label{tb:31visda}
	\addtolength{\tabcolsep}{-2pt}
	\begin{center}
	\scalebox{0.75}{
		\begin{tabular}{l|cccccc|c|c}
			\hline
			\multirow{2}{*}{Method}& \multicolumn{6}{c|}{\textbf{Office-31}}&&\textbf{VisDA}\\
			& A2D	& A2W & D2W & W2D & D2A & W2A& Avg&Per-class\\\cline{1-9}
			Source model     &74.1 & 95.3 & 99.0 & 80.1 & 54.0 & 56.3 & 76.5 &  46.3 \\
			
			CDAN~\cite{long2018conditional}& 93.1 & 98.2 & 100 & 89.8 & 70.1 & 68.0 & 86.6 & 70.0\\
			
			MDD~\cite{zhang2019bridging}& 94.5 & 98.4 & 100 & 93.5 & 74.6 & 72.2 & 74.6\\
			DMRL~\cite{wu2020dual} &93.4  &90.8  &{99.0} &100.0 &{73.0}  &71.2 &87.9  & 75.5\\
			{BNM}~\cite{cui2020towards}  &90.3  &91.5  &98.5 &{100.0} &70.9  &71.6 &87.1&-\\	
			SRDC \cite{tang2020unsupervised}&{95.8}  &{95.7}  &{99.2} &100.0 &{76.7}  &77.1 &\textbf{90.8}&-\\
			MCC~\cite{jin2019minimum}&{95.6}  &{95.4}  &98.6 &100.0 &{72.6}  &73.9 &{89.4} &78.8\\
   {DWL}~\cite{xiao2021dynamic}&{91.2}&{89.2}&{99.2}&{100.0}&{73.1}&{69.8}&{87.1}&{77.1}\\
			\hline
			3C-GAN~\cite{li2020model} & {92.7} &  93.7 &  98.5 & 99.8  & {75.3} & {77.8} & {89.6} &81.6\\
			SHOT~\cite{liang2020we}& {93.1} &  90.9 &  98.8 & 99.9  & 74.5 & 74.8 & {88.7}&82.9\\
			\textbf{Ours}&92.0	&{94.6} & 98.1  &{100.0}	&{74.6}	&{75.2}	&{89.1}&\textbf{83.0}\\
			\hline
		\end{tabular}
	}
\end{center}
\vspace{-6mm}
\end{table}

\begin{table}[t]
\caption{Results on domain adaptation on Office-Home. The two methods at the bottom are \textbf{source-free} methods.  {'Avg' means average precision over all tasks.}\vspace{-4mm}}\label{tb:oh}
\addtolength{\tabcolsep}{-4pt}
\begin{center}
\scalebox{0.70}{
	\begin{tabular}{l|cccccccccccc|c}
		\hline
		\multirow{2}{*}{Method}&\multicolumn{12}{c|}{\textbf{Office-Home}}&\\
		&  A2C & A2P & A2R&C2A&C2P & C2R & P2A & P2C & P2R & R2A&	R2C & R2P & Avg\\\cline{1-14}
		Source model     & 37.0 & 62.2 & 70.7 & 46.6 & 55.1 & 60.3 & 46.1 & 32.0 & 68.7 & 61.8 & 39.2 & 75.4 & 54.6  \\
		
		CDAN~\cite{long2018conditional}& 49.0 & 69.3 & 74.5 & 54.4 & 66 & 68.4 & 55.6 & 48.3 & 75.9 & 68.4 & 55.4 & 80.5 & 63.8\\
		
		MDD~\cite{zhang2019bridging}& 54.9 & 73.7 & 77.8 & 60.0 & 71.4 & 71.8 & 61.2 & 53.6 & 78.1 & 72.5 & 60.2 & 82.3 & 68.1 \\
		{BNM}~\cite{cui2020towards}  &52.3 & 73.9 & {80.0} & 63.3 & {72.9} & {74.9} & 61.7 & {49.5} & {79.7} & 70.5 & {53.6} & 82.2 & 67.9\\	
		SRDC \cite{tang2020unsupervised}&52.3 & 76.3 & {81.0} & {69.5} & {76.2} & {78.0} & {68.7} & {53.8} & {81.7} & {76.3} & {57.1} & {85.0} & {71.3}\\
		\hline
		SHOT~\cite{liang2020we}&  57.1&{78.1} & 81.5 & {68.0} & {78.2} & {78.1} & {67.4} & 54.9 & {82.2} & 73.3 & 58.8 & {84.3} & \textbf{71.8}\\
		\textbf{Ours}&{57.4} 	&{77.5} 	&{82.4} 	&{68.0} 	&{77.2} 	&{75.1} 	&{67.1} 	&{55.5} 	&{81.9} 	&73.9 	&{59.5} 	&{84.2} 	&{71.6}\\
		\hline
	\end{tabular}
}
\end{center}
\vspace{-4mm}
\end{table}

\subsection{Offline Source-free Domain Adaptation}\label{sec:exper_offline}

\noindent  \textbf{Datasets}. We use three benchmark datasets. \textbf{Office-31}~\cite{saenko2010adapting} contains 3 domains (Amazon  {denoted as A}, Webcam  {denoted as W}, DSLR  {denoted as D}) with 31 classes and 4,652 images. \textbf{Office-Home}~\cite{venkateswara2017deep} contains 4 domains (Real  {denoted as R}, Clipart  {denoted as C}, Art  {denoted as A}, Product  {denoted as P}) with 65 classes and a total of 15,500 images. \textbf{VisDA-2017}~\cite{peng2017visda} ({denoted as VisDA}) is a more challenging dataset, with 12-class synthesis-to-real object recognition tasks, its source domain contains 152k synthetic images while the target domain has 55k real object images.

\noindent \textbf{Model details} We adopt the backbone of ResNet-50~\cite{he2016deep} (for office datasets) or ResNet-101 (for VisDA) along with an extra fully connected (fc) layer as feature extractor, and a fc layer as classifier head. We adopt SGD with momentum 0.9 and batch size of 128 on all datasets. On the source domain, the learning of the ImageNet pretrained backbone and the newly added layers are 1e-3 and 1e-2 respectively, except for the ones on VisDA, which are 1e-4 and 1e-3 respectively. We further reduce the learning rate 10 times training on the target domain. We train 20 epochs on the source domain, and 30 epochs on the target domain.
All experiments are conducted on a single RTX 6000 GPU. All results are reported from the classifier $C_1$, and are the average across three running with random seeds.

\noindent  \textbf{Quantitative Results.} The results under offline setting on the three datasets are shown in Tab.~\ref{tb:31visda} and Tab.~\ref{tb:oh}. In these tables, the top part shows results for the normal setting with access to source data during adaptation. The bottom one shows results for the source-free setting. 
As reported in Tab.~\ref{tb:31visda} and Tab.~\ref{tb:oh}, our method outperforms most methods which have access to source data on all these datasets. 

The proposed method still obtains the comparative performance when comparing with current source-free methods. In particular, our method surpasses SHOT~\cite{liang2020we} by 0.1\%, and 3C-GAN~\cite{li2020model} by 1.4\% on the more challenging VisDA dataset (Tab.~\ref{tb:31visda}), and gets closer results on Office-Home (Tab.~\ref{tb:oh}) compared to SHOT. Note that 3C-GAN highly relies on the extra synthesized data. On Office-31 (Tab.~\ref{tb:31visda}), the proposed BAIT achieves better result than SHOT, and competitive result to 3C-GAN. The reported results clearly demonstrate the efficacy of the proposed method without access to the source data during adaptation.

\begin{table}[t]
\centering
\caption{(\textbf{Left}) Ablation study. $fix$ means fixing the first classifier, $spl$ means entropy splitting, $\mathcal{L}_b$ is the class-balance loss. (\textbf{Right}) Ablation study on $\tau$. All  {uncertain} means no splitting actually.}\label{tab:aba_mcd} \label{tab:aba_tau}
\begin{minipage}[tbp]{0.2\textwidth}
\addtolength{\tabcolsep}{-3pt}
	\scalebox{0.9}{\begin{tabular}{ccccc}
			\hline
			SF&$fix$ &$spl$ &$\mathcal{L}_b$ &Avg.\\
			\hline
			$\times$&$\times$&$\times$&$ \times$&{65.3}\\
			$\surd$&$\times$&$\times$&$\times$&{60.8} \\
			$\surd$&$\surd$&$\times$&$ \times $&{67.3}\\
   $\surd$&$\times$&$\surd$&$ \times $&{60.2}\\
   $\surd$&$\times$&$\times$&$ \surd $&{52.7}\\
			$\surd$&$\surd$&$\surd$&$\times $&{68.2}\\
			$\surd$&$\surd$&$\surd$&$ \surd $&\textbf{71.6}\\
			\hline
\end{tabular}}
\end{minipage}
\begin{minipage}[tbp]{0.25\textwidth}
	\addtolength{\tabcolsep}{-4pt}
	\scalebox{0.8}{\begin{tabular}{cc}
			\hline
			\textbf{Office-Home}&Avg.\\
			\hline
			BAIT ($\tau$ as all  {uncertain}) &{70.8} \\
			BAIT ($\tau$ as 75\%  {uncertain}) &{71.2} \\
			BAIT ($\tau$ as 50\%  {uncertain}, in paper)&\textbf{71.6} \\
			BAIT ($\tau$ as 25\%  {uncertain})&{70.2} \\
			\hline
	\end{tabular}}
\end{minipage}
\vspace{-4mm}
\end{table}

\begin{table*}[t]
\caption{(\textbf{Top}) Results on \textbf{online} source-free domain adaptation on Office-31 and Office-Home.  {'Avg' means average precision over all tasks. The three methods at the bottom are \textbf{source-free} methods.} (\textbf{Bottom}) {Results on \textbf{online} source-free domain adaptation on VisDA.  'Per-class' means average per-class accuracy over 12 classes. The three methods at the bottom are \textbf{source-free} methods.}}\label{tb:online} \label{tb:vis_online}
\begin{minipage}[tbp]{0.95\textwidth}
\centering
\addtolength{\tabcolsep}{-3pt}
\begin{center}
\scalebox{0.75}{
\begin{tabular}{l|cccccc|c|cccccccccccc|c}
	\hline
	\multirow{2}{*}{Method}& \multicolumn{6}{c|}{\textbf{Office-31}}&&\multicolumn{12}{c|}{\textbf{Office-Home}}&\\
	& A2D	& A2W & D2W & W2D & D2A & W2A& Avg& A2C & A2P & A2R&C2A&C2P & C2R & P2A & P2C & P2R & R2A&	R2C & R2P & Avg\\
	\hline
	MCD (\textit{w/} source)~\cite{saito2018maximum} &87.9	&86.1&	92.3&	96.9&	62.2&	65.3	&81.8 &
	46.3	&65.3	&74.9	&57.2&	64.3&	66.0	&55.1&	45.3&	75.1&	66.7&	48.4&	78.1	&61.9 \\
	
	\hline
	MCD (\textit{w/o} source)~\cite{saito2018maximum} & 85.8	&82.4	&91.0	&96.0	&59.6	&62.3	&79.5&  44.2& 63.7& 72.7& 50.6& 63.9& 60.2& 54.8& 42.4& 73.0& 60.2& 47.5& 76.9&59.2\\
	
	SHOT~\cite{liang2020we} & {87.8} &  85.2 &  96.0 & 99.6  & 70.1 & 68.3 & {84.6}& 48.1&{72.0} & 76.2 & {59.0} & {68.9} & {67.8} & {58.7} & 47.0 & {77.3} & 70.0 & 53.8 & {80.1} &  {{64.9}}\\
	
	\textbf{BAIT (ours)} &90.8&{86.2} & 96.5  &{99.8}	&{71.5}	&{69.6}	&\textbf{85.7}&{51.0} 	&{72.9} 	&{77.4} 	&{60.9} 	&{71.0} 	&{68.7} 	&{60.7} 	&{49.3} 	&{78.2} 	&70.2 	&{54.5} 	&{80.4} 	& {\textbf{66.3}} \\
	
	\hline
\end{tabular}
}
\end{center}
\vspace{-2mm}
\end{minipage}
\begin{minipage}[tbp]{0.9\textwidth}
\centering
\addtolength{\tabcolsep}{-4pt}
\begin{center}
\vspace{1mm}
\scalebox{0.9}{
\begin{tabular}{l|cccccccccccc|c}
\hline
Method & plane & bcycl & bus & car & horse & knife & mcycl & person & plant & sktbrd & train & truck & Per-class \\
\hline
MCD (\textit{w/} source) &77.8&	53.1&	79.6&	55.3&	80.1&	64.3&	78.9&	62.6&	68.7&	40.5&	77.2&	20.4
&63.2\\	\hline
MCD (\textit{w/o} source)& 71.4&	47.7&	74.7&	50.2	&76.8&	62.3&	73.1&	60.7&	65.4&	37.5&	73.2&	17.6&
59.2\\

SHOT & 89.5&	59.9&	85.6&	57.3&	87.6&	72.4&	89.3&	70.0&	75.1&	45.3&	82.4&39.8
&71.2\\
\textbf{BAIT (ours)}& 93.2&	66.2&	87.1&	59.1&	90.2&	76.9&	92.1&	83.4&	80.6&	49.5&	87.7&	45.7
&\textbf{76.0}\\
\hline
\end{tabular}
}
\end{center}
\end{minipage}
\end{table*}

\begin{figure}[btp]
\begin{center}
\includegraphics[width=0.51\textwidth]{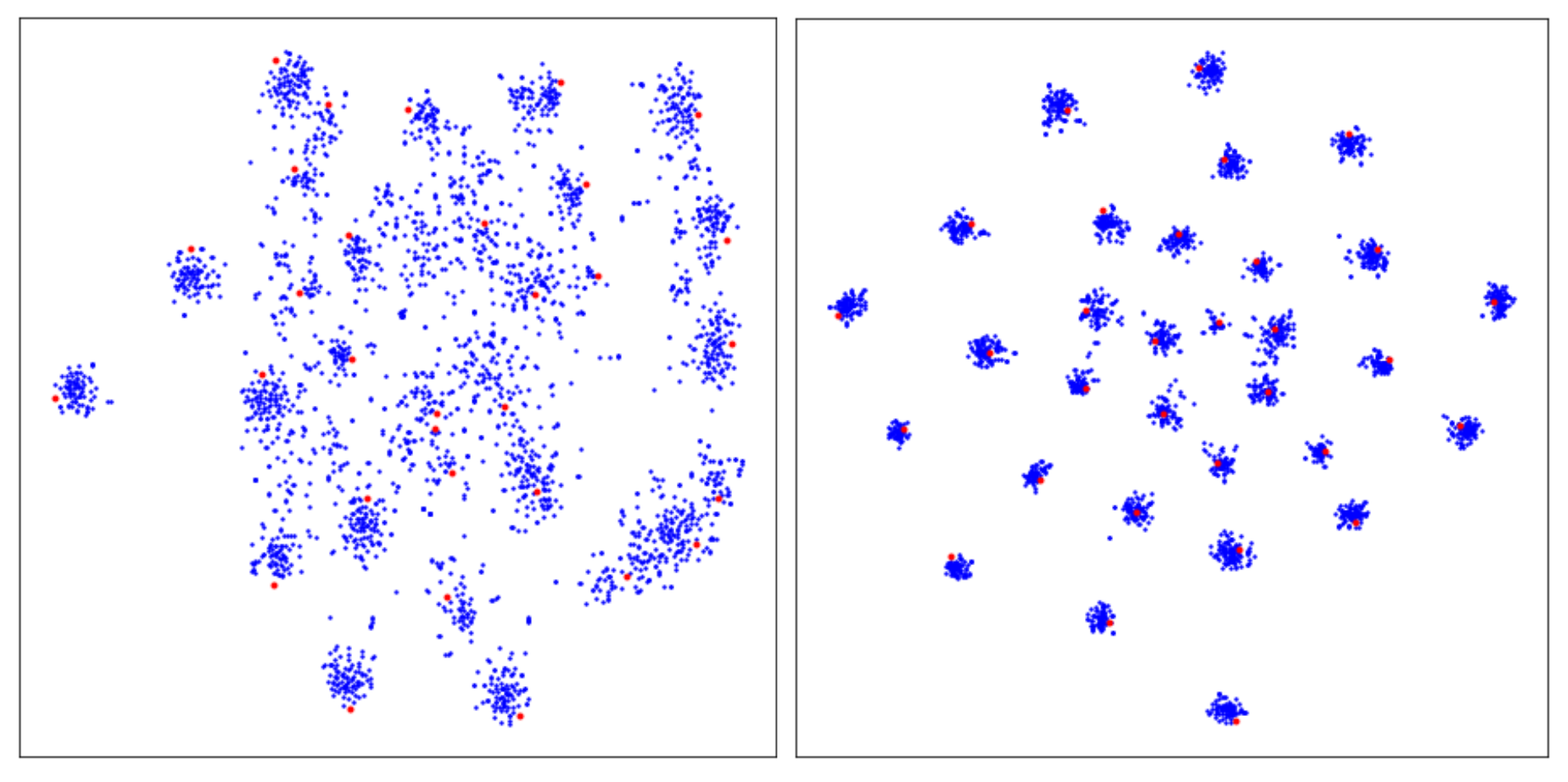}
\end{center}
\vspace{-4mm}
\caption{{t-SNE visualization (Left: Source only; Right: BAIT) for features from the target domain on W$->$A (Office-31). The red points are the class prototype from $C_1$. Best seen in screen.\vspace{-4mm}}}
\label{fig:cm_tsne}
\vspace{-2mm}
\end{figure}

\noindent \textbf{Ablation Study.} We conduct a detailed ablation study to isolate the performance gain due to key components of our method. Note the existing domain adaptation datasets do not provide train/validation/test splitting, here we directly conduct the ablation study on test set, just as all existing methods did. We start from a variant of MCD which is reproduced by ourselves as a baseline (the first and second row in Tab.~\ref{tab:aba_mcd}), note we replace the L1 distance in the original MCD with Eq.\ref{eq:wandering_set} and Eq.\ref{eq:casting_bait} used in our paper. As shown by the results in Tab.~\ref{tab:aba_mcd} on the Office-Home, if removing the access to source data ($SF$), significant degrading will occur for MCD. Then with our proposed modules on top of this baseline: fixing the first classifier ($fix$), entropy splitting ($spl$) and class-balance loss ($\mathcal{L}_b$), it performs well under SFDA setting. The experimental results show the effectiveness of the proposed method and the importance of all components.  {In addition, we ablate $\mathcal{L}_{cast}$ which is used to train the auxiliary classifier, and $\mathcal{L}_{bite}$ which trains the feature extractor. Both are necessary components of our method, removing any one of these results in very bad performance: removing $L_{cast}$ obtains 45\% and removing $\mathcal{L}_{bite}$ gets only 8\%.}

We also report results with different $\tau$. 
In all experiments, we have set $\tau$ as to select half of the \textit{current batch} as certain and uncertain set. 
Here, in Tab.~\ref{tab:aba_tau}, we also choose $\tau$ to select 100\%, 75\% or 25\% of the \textit{current batch} as the  {uncertain} set, the results show our method is not very sensitive to the choice of this hyperparameter, we posit that this is because of the random mini-batch sampling where the same image can be grouped into both certain and uncertain set in different batches during training.

\begin{figure}[!tbp]
\begin{center}
\includegraphics[width=0.49\textwidth]{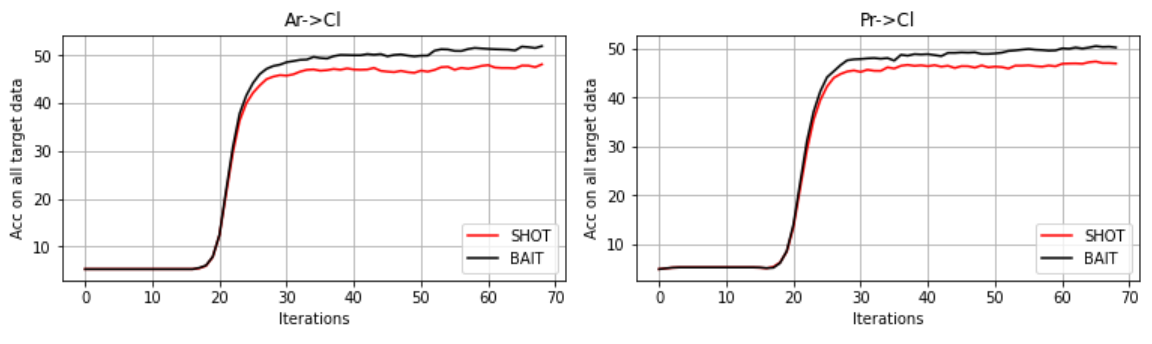}
\end{center}
\vspace{-4mm}
\caption{{Accuracy on the whole target data during online adaptation on Office-Home.\vspace{-4mm}}}
\label{fig:acc_online}
\vspace{-2mm}
\end{figure}

\noindent \textbf{Embedding visualization.} Fig.~\ref{fig:cm_tsne} shows the t-SNE visualization of target features obtained with the source model and after adaptation with BAIT. Target features form more compact and clear clusters after BAIT than in the source model, indicating that BAIT produces more discriminative features. We also show the class prototypes (red points) which are the weights of classifier $C_1$, it shows target features cluster around the corresponding prototypes.

\subsection{Online Source-free Domain Adaptation}

We also report results for the online setting, where all target data can only be accessed once, \textit{i.e.}, training for only one epoch. All datasets and model details stay the same as in the offline setting in Sec.~\ref{sec:exper_offline}. After one epoch training, we evaluate the model on the target data. This setting is important for some online streaming situations, where the system is expected to directly process the target data and data cannot be revisited.
Note under this setting we abandon the entropy splitting. We reproduce SHOT~\cite{liang2020we} under this setting as the authors released their code. As shown in Tables~\ref{tb:vis_online}, our BAIT outperforms SHOT on all three datasets. {The accuracy curve for two tasks of Office-Home during adaptation is shown in Fig.~\ref{fig:acc_online}. They show our method is outperforms SHOT. The reason of lower starting accuracy is because we initialize all BN layers with zero means and variance equal to one before adaptation, which was found to lead to better results}. {Note here the model cannot access all target data in each mini-batch training, thus SHOT can only use the current mini-batch to compute pseudo labels. This means that the computed pseudo labels are quite similar with the naive pseudo label from the model, thereby compromising the performance. This is the reason SHOT gets lower results than BAIT.} {BAIT is an extension of MCD. Tables~\ref{tb:online} shows that indeed the proposed changes do considerably impact performance and our method without source data even outperforms MCD with source data.}

\section{Conclusion}
There are many practical scenarios where source data may not be available (e.g. due to privacy or availability restrictions) or may be expensive to process. In this paper we study this challenging yet promising domain adaptation setting (i.e. SFDA), and propose BAIT, a fast and effective approach. BAIT aligns target features with fixed source classifier via an extra bait classifier that locates uncertain target features and drags them towards the right side of the source decision boundary. The experimental results show that BAIT achieves competitive performance on several datasets under the offline setting, and  surpasses other SFDA methods in the more realistic online setting.

\paragraph{\textbf{Acknowledgement}}
{
We acknowledge the support from Huawei Kirin Solution, and the project PID2019-104174GB-I00 (MINECO, Spain), TED2021-132513B-I00, and PID2021-128178OB-I00 (MICINN, Spain), and the Ramón y Cajal fellowship RYC2019-027020-I, and the CERCA Programme of Generalitat de Catalunya.
}

\bibliographystyle{elsarticle-num}
\bibliography{refs}


\end{document}